\documentclass[letterpaper, 10 pt, conference]{ieeeconf}  

\IEEEoverridecommandlockouts                              

\usepackage{hyperref}
\usepackage[latin9]{inputenc}
\usepackage{amsmath}
\usepackage{amssymb}
\usepackage{graphicx}
\usepackage{float}
\usepackage{enumerate}
\usepackage{tikz}
\usetikzlibrary{shapes,arrows,automata,calc,trees,positioning,fit,shapes,calc,patterns,3d}
\usepackage{pgfplots}
\usepgfplotslibrary{fillbetween}
\usepackage{booktabs}
\usepackage{xparse}
\usepackage{algorithm,algorithmic}
\usepackage{subfigure}
\usepackage[normalem]{ulem}
\allowdisplaybreaks

\newcommand{\mc}[1]{\mathcal{#1}}
\newcommand{\bs}[1]{\boldsymbol{#1}}
\newcommand{\routes}{\boldsymbol{\alpha}}
\newcommand{\routingvars}{\alpha_{ij}^k}
\newcommand{\tc}{\boldsymbol{x}}    
\newcommand{\ttc}{\boldsymbol{x}_T} 
\newcommand{\ntc}{\boldsymbol{x}_N} 

\DeclareMathOperator{\erf}{erf}

\usepackage{cite}
\usepackage{url}

\title{\LARGE \bf
Mobile Wireless Network Infrastructure on Demand
}

\author{Daniel Mox$^{1}$, Miguel Calvo-Fullana$^{1}$, Mikhail Gerasimenko$^{2}$\\ Jonathan Fink$^{3}$, Vijay Kumar$^{1}$ and Alejandro Ribeiro$^{1}$
\thanks{We gratefully acknowledge the support of ARL grant DCIST CRA W911NF-17-2-0181, NSF Grants CNS-1446592, and CNS-1521617, ARO grant W911NF-13-1-0350, Qualcomm Research, United Technologies, the Intel Science and Technology Center for Wireless Autonomous Systems, and National Science Foundation Graduate Research Fellowship Grant No. DGE-1845298.}%
\thanks{$^{1}$Daniel Mox, Miguel Calvo-Fullana, Vijay Kumar, and Alejandro Riberio are with the University of Pennsylvania, Philadelphia, PA, USA}
\thanks{$^{2}$Mikhail Gerasimenko is with Tampere University, Tampere, Finland}%
\thanks{$^{3}$Jonathan Fink is is with the U.S. Army Research Laboratory, Adelphi, MD, USA}
}

\begin{document}

\maketitle
\thispagestyle{empty}
\pagestyle{empty}

\begin{abstract}

In this work, we introduce \emph{Mobile Wireless Infrastructure on Demand}: a framework for providing wireless connectivity to multi-robot teams via autonomously reconfiguring ad-hoc networks. In many cases, previous multi-agent systems either assumed the availability of existing communication infrastructure or were required to create a network in addition to completing their objective. Instead our system explicitly assumes the responsibility of creating and sustaining a wireless network capable of satisfying end-to-end communication requirements of a team of agents, called the task team, performing an arbitrary objective. To accomplish this goal, we propose a joint optimization framework that alternates between finding optimal network routes to support data flows between the task agents and improving the performance of the network by repositioning a collection of mobile relay nodes referred to as the network team. We demonstrate our approach with simulations and experiments wherein wireless connectivity is provided to patrolling task agents.

\end{abstract}

\section{Introduction}

Wireless networks have become a ubiquitous part of modern life. Today, technologies such as Wi-Fi and cellular networks blanket vast portions of the globe enabling unprecedented access to information. As wireless communication becomes more tightly integrated into everyday life it is increasingly normal to presume on it's availability and performance. However, providing widespread wireless connectivity at scale requires tremendous effort both in coordinating large distributed systems and deploying and maintaining costly infrastructure.

While areas not covered by fixed infrastructure continue to shrink, there still remain cases where existing wireless networks are unavailable such as underground environments, rural regions, and areas affected by natural disaster. In these scenarios, reliance on existing technology is not possible and new approaches to providing connectivity must be explored,  particularly as it relates to fast deployment of wireless infrastructure.
 
With these challenges in mind we introduce a system for delivering \emph{Mobile Wireless Infrastructure on Demand}. Our approach leverages mobile robots equipped with wireless hardware, referred to as the network team, to create and sustain communication networks in dynamic environments. Client users, referred to collectively as the task team, seeking to accomplish objectives requiring communication can connect to the provided network and the network nodes reconfigure and route information to satisfy their demands. While our philosophy would extend to any autonomous robotic platform, due to their versatility, we consider Unmanned Aerial Vehicles (UAVs) equipped with IEEE 802.11 communication hardware which act as an ad-hoc network.

A considerable amount of research has been devoted to studying the effect communication links and wireless networks have on the ability of agents to coordinate. One common approach leverages concepts from algebraic graph theory to maximize or preserve the algebraic connectivity of a state dependent Laplacinan \cite{kim2005maximizing, stump2008connectivity, zavlanos2007potential, zavlanos2008distributed, ji2007distributed, de2006decentralized} which can be solved in both a centralized \cite{kim2005maximizing, stump2008connectivity, zavlanos2007potential} and decentralized manner \cite{zavlanos2008distributed, de2006decentralized, ji2007distributed}. Wireless channels themselves are extremely difficult to predict and approaches can be categorized by the abstraction they use to model point-to-point communication. Disk models are the simplest and consider all agents within some distance to be in communication range \cite{ji2007distributed, zavlanos2008distributed, spanos2005motion, notarstefano2006maintaining}. Other approaches employ a function of the inter-robot distance, often exponential decay, to model the channel \cite{de2006decentralized, schuresko2009distributed, zavlanos2012network, zavlanos2007potential, kim2005maximizing, stump2008connectivity, tekdas2010robotic}. Probabilistic models seek to honor the inherent uncertainty associated with predicting wireless channels by provided both an expected channel rate and corresponding variance \cite{mostofi2008communication, mostofi2009characterization, yan2012robotic, fink2013robust, fink2011communication}.


In this work we take a similar approach to \cite{fink2011robust} and more recently \cite{stephan2017concurrent}. Both utilize a robust routing communication protocol, which seeks to overcome channel uncertainty using link redundancy, to judge the \emph{feasibility} of network configurations during planning. In this work we take a different approach by directly seeking configurations which \emph{maximize} network performance. Additionally, both works abstract the task as a potential function that must be driven to zero by a single team of agents also responsible for satisfying the associated communication requirements. Instead, we explicitly separating agents into task and network teams. This has multiple benefits. First, it decouples network considerations from task planning, which is often a challenging problem by itself. Secondly, it allows our approach to remain task agnostic. While many robotic tasks can be reduced to potential functions (e.g., visiting a goal location) others cannot or must be reasoned about in a more abstract manner. Instead of making limiting assumptions about the type of objective being performed, we define an interface where end-to-end communication rate requirements are provided to the task team by the network team. In this way our system can be applied to any multi-agent system that can specify its communication requirements.

The rest of this paper is organized as follows: in Section \ref{sec:methodology} we present our methodology, divided into channel modeling (\ref{sec:channel_model}), network flows (\ref{sec:network_flows}), probabilistic routing (\ref{sec:prob_routing}), and network reconfiguration (\ref{sec:network_reconfig}), in Section \ref{sec:num_results} we discuss our simulations and experimental results, and finally in Section \ref{sec:conclusions} we provide concluding remarks.

\section{Problem Introduction and Methodology}\label{sec:methodology}

\begin{figure}[t]
    \centering
  	\includegraphics[scale=1]{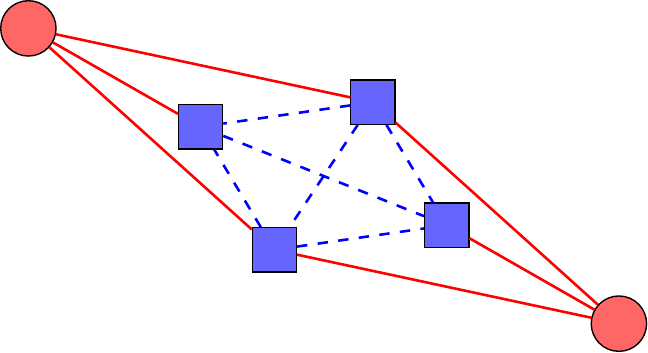}
    \caption{An example network configuration with task agents as red circles, network agents as blue squares, task-network connections as solid red lines, and inter-network connections as dashed blue lines.}
    \label{fig:example_network}
\end{figure}

Consider Fig. \ref{fig:example_network}, consisting of a team of mobile robots collaborating to complete a task requiring communication. Instead of creating and sustaining a wireless network in addition to completing their objective, this task team presumes on the ability to communicate while a different group of robots, called the network team, provides the required infrastructure. The network team positions itself in the environment and routes packets such that the task team can go about their objective without concern for the impact of their actions on their ability to communicate. The focus of this paper is on determining network team configuration and routing variables to support the task team's presumption.

Before proceeding, we introduce some common notation. Suppose there are $p$ agents in the task team and $q$ agents in network team with $n = p+q$ total agents. $\mc{I}_T$ and $\mc{I}_N$ represent the ordered set of indices of the task agents and network agents so that $|\mc{I}_T|=p$, $|\mc{I}_N|=q$, $\mc{I}_T \cup \mc{I}_N = \left\{1,2,\ldots,n\right\}$, and $\mc{I}_T \cap \mc{I}_N = \emptyset$ (i.e. dual citizenship is not permitted). The position of each agent is given by $x_i \in \mathbb{R}^2$ and the state of the combined teams is $\tc=\left[ x_1, x_2, \ldots, x_n \right ]^T$. The configuration of the task team is $\bs{x}_T=\left[x_{\mc{I}_T(1)}, \ldots, x_{\mc{I}_T(p)} \right]^T$ and the network team is $\ntc=\left[x_{\mc{I}_N(1)}, \ldots, x_{\mc{I}_N(q)} \right]^T$.

\subsection{Channel Model}\label{sec:channel_model}

\begin{figure}[t]
	\centering
	\includegraphics[scale=1]{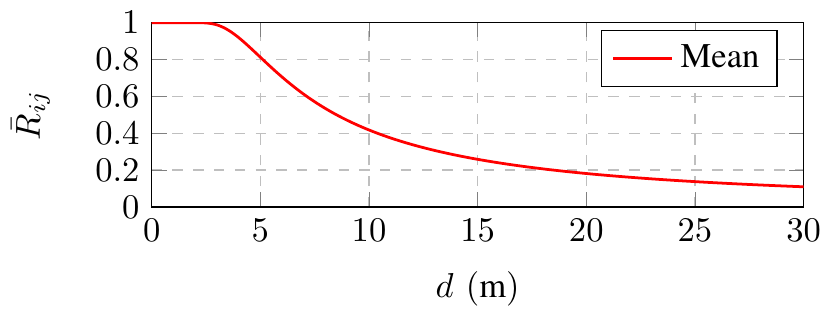} \\
	\includegraphics[scale=1]{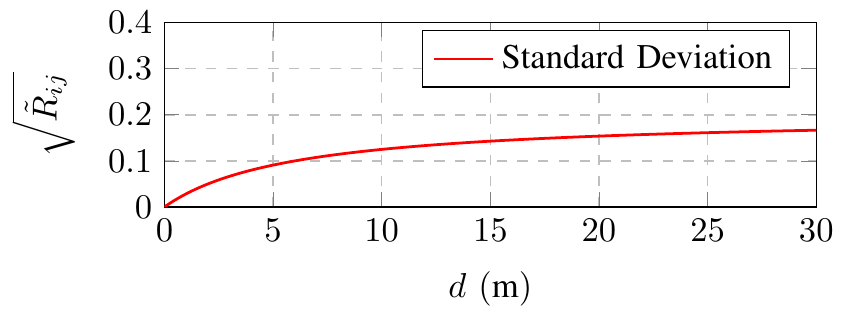}
	\caption{Characterization of the expected rate function $\bar{R}(x_i,x_j)$ in equation \eqref{eq:channel_mean} and variance $\tilde{R}(x_i,x_j)$ in equation \eqref{eq:channel_var}, where $d=\|x_i-x_j\|$ is the distance between the agents $x_i$ and $x_j$, and $P_{L_0}=-53~\text{dBm}$, $n=2.52$, $P_{N_0}=-70~\text{dBm}$, $a=0.2$ and $b=0.6$.}
	\label{fig:rate}
\end{figure}

In order to reason about the rate at which packets can be transmitted in a network, it is necessary to predict the state of wireless links between pairs of agents. In complex real-world scenarios, this is in itself a challenging task with a large body of research focused on establishing empirical models across a wide range of hardware and environments. In this work, we consider a model that seeks to 1) capture the dominant characteristics of wireless channels without incurring significant model complexity and 2) remain somewhat hardware agnostic. In general, as the distance between two agents grows one can expect their communication rate to decrease and the uncertainty associated with that estimate to increase. This perspective emits a probabilistic approach to channel rate modeling.

More formally, we consider a function $R_{ij}(x_i,x_j) \in [0,1]$ representing a random variable of the normalized available transmission rate of a point-to-point link. Such a value depends on the randomness of the communication channel which we assume can be described by an underlying distribution characterized by its mean and variance, $\bar{R}_{ij}$ and $\tilde{R}_{ij}$, respectively. We consider the following characterization, which is also plotted in Figure \ref{fig:rate}:
\begin{align}
    \bar{R}_{ij} = \bar{R}(x_i,x_j) = \erf \left( \sqrt{\frac{P_{L_0}}{P_{N_0}}||x_i-x_j||^{-n}} \right),
    \label{eq:channel_mean}
\end{align}
\begin{align}
    \tilde{R}_{ij} = \tilde{R}(x_i,x_j) = \frac{a||x_i-x_j||}{b+||x_i-x_j||},
    \label{eq:channel_var}
\end{align}
where $\erf(x)=\frac{1}{\sqrt{\pi}} \int_{-x}^{x} e^{-t^2}dt$ is the Gauss error function, $P_{L_0}$ is the transmit power, $n$ is the decay rate, $P_{N_0}$ is the noise at the receiver, and $a$ and $b$ are decay uncertainty parameters. This model is formed by composing generic models for received signal strength and bit error rate to obtain a channel rate estimate \cite{shorey2006mobile}. Note that this model can be used for a wide range of hardware by adjusting these parameters accordingly. Furthermore, it can also accommodate noisy, cluttered environments (e.g. indoor) where reliable estimates are scarce by adjusting the uncertainty parameters $a$ and $b$.

\subsection{Network Flows}\label{sec:network_flows}

As the task team moves to accomplish its objective, data is transmitted between source node(s) and destination node(s) in the task team via the network team. These flows of information are indexed by $k\in\left\{ 1,2,\ldots,K \right\}$ with the set of source and destination nodes for flow $k$ given by $S_k, D_k\in\mc{I}_T$, respectively. As an example, a foraging agent $i$ ($S_k=\{i\}$) in a collaborative mapping task might transmit newly gathered observations of the environment to the rest of the team ($D_k=\mc{I}_T \backslash S_k$). Each flow requires a minimum communication rate in order to be transmitted successfully. The required minimum rate at source node $i$ for flow $k$ is given by $m_i^k$. For network team nodes that do not contribute new packets to the network but simply act as relays, $m_i^k=0$.

Data packets are relayed through the network according to routing variables $\routingvars \in [0,1]$ which represent the probability or fraction of an arbitrary time step node $i$ spends transmitting data associated with flow $k$ to node $j$. The set of all routing variables is denoted by $\routes$. Routing variables must satisfy $\sum_{j,k}\routingvars \leq 1$ as well as $\sum_{i,k}\routingvars \leq 1$, which are the upper bound on channel transmission and receiver usage during a time step. Finally, task agents access the network as clients; if task agent $i$ is the source of flow $k$ then $\routingvars = 0 \ \forall j$ (i.e. data is not returned to the source node) and similarly if agent $j$ is the destination then $\routingvars=0 \ \forall i$ (i.e., destinations nodes don't rebroadcast data once it has been received). The difference between data transmitted and received at a node, called the rate margin, is given by:
\begin{align}
  b_i^k(\routes, \tc) = \sum_{j=1}^n\routingvars R_{ij} - \sum_{j=1}^n\alpha_{ji}^k R_{ji},
  \label{eq:rate_margin_flow}
\end{align}
and since the rates $R_{ij}=R(x_i,x_j)$ are random variables, $b_i^k$ is also a random variable:
\begin{align}
  \bar{b}_i^k(\routes, \tc) = \sum_{j=1}^n\routingvars\bar{R}_{ij} - \sum_{j=1}^n\alpha_{ji}^k\bar{R}_{ji},
  \label{eq:expected_rate_margin}
\end{align}
\begin{align}
  \tilde{b}_i^k(\routes, \tc) = \sum_{j=1}^n(\routingvars)^2\tilde{R}_{ij} + \sum_{j=1}^n(\alpha_{ji}^k)^2\tilde{R}_{ji},
    \label{eq:var_rate_margin}
\end{align}
where $\bar{b}_i^k$ is the expected rate margin of node $i$ for flow $k$, and $\tilde{b}_i^k$ is the confidence in the expected rate margin. In order for packets to be routed through the network and delivered as required it is sufficient to ensure the flow of data into and out of each node remains balanced, preventing unbounded accumulation of packets at any node:
\begin{align}
    \bar{b}_i^k \geq m_i^k.
    \label{eq:deterministic_constraint}
\end{align}

\subsection{Probabilistic Routing}\label{sec:prob_routing}

Since the rate margin $b_i^k$ is a random variable [cf. expression \eqref{eq:rate_margin_flow}], equation \eqref{eq:deterministic_constraint} can only be satisfied in a probabilistic sense. Namely,
\begin{equation}
  \mathbb{P}\left[ \bar{b}_i^k(\routes, \tc) \ge m_i^k \right] \geq 1-\epsilon_k,
  \label{eq:probability_constraint}
\end{equation}
where $\epsilon$ is the risk of the constraint being unsatisfied and $1-\epsilon$ is the confidence with which the constraint is met. Thus, the pair $(m_i^k,1-\epsilon_k)$ forms the communication rate specification requested by the user. Specifically, the $i$-th user demands for the $k$-th flow an expected communication rate of $m_i^k$ with confidence $1-\epsilon_k$.

For a wide class of probability distributions expression \eqref{eq:probability_constraint} can be satisfied using Chebyshev's inequality. However, tighter bounds can be achieved if the distribution is known. In the following, we assume that the rate margin $b_i^k$ is normally distributed and equation \eqref{eq:var_rate_margin} can then be expressed as
\begin{equation}
    \frac{\bar{b}_i^k(\routes,\tc) - m_i^k}{\sqrt{\tilde{b}_i^k(\routes,\tc)}} \geq \Phi^{-1}(1-\epsilon_k),
    \label{eq:socp_probability_constraint}
\end{equation}
where $\Phi^{-1}(\cdot)$ is the inverse normal cumulative distribution function. Now, we intend to find, given team configuration $\bs{x}$, the set of routing variables $\routes$ which satisfy equation \ref{eq:socp_probability_constraint}. There may be many sets of routing variables $\routes$ that satisfy equation \eqref{eq:socp_probability_constraint}; we are interested in those that do so with the greatest margin. Thus, a non-negative slack variable $s$ is introduced and the robust routing problem is posed as follows
\begin{subequations}
  \begin{align}
    \underset{\routes \in \mathcal{A},s \geq 0}{\text{maximize}} \quad& s\\
    \text{subject to} \quad& \frac{\bar{b}_i^k(\routes,\tc) - m_i^k -s}{\sqrt{\tilde{b}_i^k(\routes,\tc)}} \geq \Phi^{-1}(1-\epsilon_k)
  \end{align}
  \label{eq:robust_routing}
\end{subequations}%
for all $i$ and $k$, where $\mathcal{A}$ is the set of implicit routing constraints composed of $\routingvars \in [0,1]$ (routing variables are a probability), $\sum_{j,k}\routingvars \leq 1$ (bounded transmission usage), $\sum_{i,k}\routingvars \leq 1$ (bounded receiver usage),  $\routingvars = 0$ for $i \in \mc{I}_T \backslash S_k$ (destination nodes don't transmit), $j \in \mc{I}_T \backslash D_k$ (packets are not returned to the source node). Following the approach in \cite{fink2011robust}, the optimization problem \eqref{eq:robust_routing} can be posed as a Second Order Cone Problem (SOCP) and a solution obtained using an available convex solver. Note that the solution to the optimization problem \eqref{eq:robust_routing} supplies valid robust routing variables given a feasible team state $\bs{x}$; in other words it says nothing about how the network agents $\ntc$ should be positioned.

\begin{figure}[t]
    \centering
    \subfigure[Low margin, high confidence ($m_i^k=0.1$,$\epsilon_k=1-0.95$).]{
  	\includegraphics[scale=1]{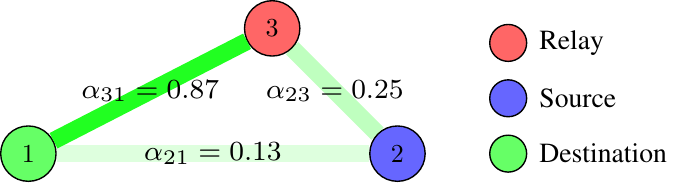}
  		\label{fig:exampleLowMarginHighConf}
	 }
    \subfigure[High margin, low confidence ($m_i^k=0.3$,$\epsilon_k=1-0.7$).]{
    \includegraphics[scale=1]{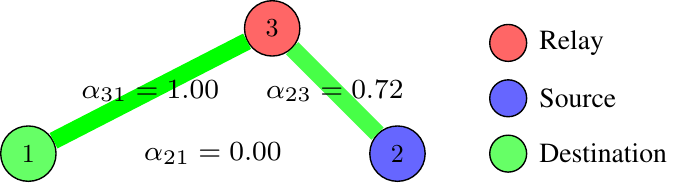}
  		\label{fig:exampleHighMarginLowConf}
	 }	 
	\caption{Optimal solutions of the robust routing problem for different margin and confidence requirements. High margin with low confidence prioritizes the strongest links while low margin with high confidence splits traffic among the routes.}
    \label{fig:exampleMarginConf}
\end{figure}

Increasing the slack $s$ is equivalent to increasing the margin with which the probability constraints in equation \eqref{eq:probability_constraint} are satisfied. This can be accomplished by increasing the expected value $\bar{b}_i^k$, by prioritizing channels with high expected rate $\bar{R}_{ij}$, and/or decreasing the variance $\tilde{b}_i^k$, by splitting routes between multiple nodes. For flows with high requirement $m_i^k$ and low confidence $1-\epsilon_k$, packets flow mainly through high rate channels; for flows with low requirement and high confidence, packets are split along multiple routes (This is illustrated in Figure \ref{fig:exampleMarginConf}). Thus, the solution to the robust routing problem \eqref{eq:robust_routing} prioritizes end-to-end performance while seeking to be robust to point-to-point failures, accounting for model uncertainty.

\subsection{Network Reconfiguration}\label{sec:network_reconfig}

Solving the robust routing problem \eqref{eq:robust_routing} allows us to find the optimal routing variables $\routingvars$ given a network team configuration $\ntc$ and task team configuration $\ttc$. Now, we are interested in finding the network team configurations $\ntc$ that satisfy the constraints of problem \eqref{eq:robust_routing} (recall only the positions of the network agents can be controlled). As the task agents move, the network agents must also adjust their positions to ensure that the communication requirements are met. Unfortunately, the rate margins in equation \eqref{eq:expected_rate_margin} are non-convex in $\tc$, precluding some kind of joint optimization of $\routes$ and $\ntc$ directly. While the gradient of slack $s$ with respect to the team configuration could be computed from the rate estimation function presented in Section \ref{sec:channel_model}, we are interested in developing a system that works for a family of rate functions meeting the requirements outlined in Section \ref{sec:channel_model}. In other words, we assume that a rate function can be efficiently queried for $\bar{R}_{ij}$,$\tilde{R}_{ij}$ but do not make further restrictions to its form (e.g., that it be differentiable).

Similar approaches solve this problem by using a sampling based gradient approximation to drive the team away from configurations that violate the communication requirements all while minimizing a task potential function \cite{fink2011robust}. In our case, the network team's singular focus is to improve the degree to which the node margin constraints of equations \eqref{eq:expected_rate_margin} are met (i.e. adjusting $\ntc$ to further increase the slack $s$ in the robust routing problem \eqref{eq:robust_routing}). While one method for doing so is to follow an approximate gradient, a more direct approach is to leverage the same samples to search for configurations that improve on the solution of \eqref{eq:robust_routing}.

Given a solution to the robust routing problem, the constraint closest to being violated can be computed from equation \eqref{eq:socp_probability_constraint} by selecting the minimum value
\begin{equation}
    \nu(\routes,\tc) = \min_{i,k} \left[ \dfrac{\bar{b}_i^k(\routes,\tc) - m_i^k}{\sqrt{\tilde{b}_i^k(\routes, \tc)}} - \Phi^{-1}(1-\epsilon_k) \right],
    \label{eq:min_constraint}
\end{equation}
where $\nu(\routes,\tc)\geq 0$ for all feasible configurations. Furthermore, the greater $\nu(\routes,\tc)$ becomes, the more room the slack $s$ has to grow in \eqref{eq:robust_routing}. Assuming solutions to the optimization problem \eqref{eq:robust_routing} do not change significantly over short distances, equation \eqref{eq:min_constraint} provides a method for checking if a neighboring location results in a better network configuration (one with higher slack) much more efficiently than resolving the SOCP. This insight forms the foundation of the local control scheme outlined in Algorithm \ref{alg:network_improvement}.

\begin{algorithm}[t]
\caption{Local controller}
\label{alg:network_improvement}
\begin{algorithmic}[1]
\renewcommand{\algorithmicrequire}{\textbf{Input:}}
\renewcommand{\algorithmicensure}{\textbf{Output:}}
\REQUIRE $\ntc, \ttc, \ntc^*$
\ENSURE  $\ntc^*$
\STATE $\routes = \texttt{SOCP}([\ttc,\ntc])$
\STATE $\routes^* = \texttt{SOCP}([\ttc,\ntc^*])$
\STATE $v^* = \nu(\routes^*,[\ttc,\ntc^*])$
\FOR {$i=0\text{ to }\texttt{max\_it}$}
\STATE $x_p = \texttt{draw\_sample}(\ntc)$
\STATE $v_p = \nu(\routes,[\ttc,\bs{x}_p])$
\IF {$v_p > v^*$}
\STATE $\ntc^* = \bs{x}_p$
\ENDIF
\ENDFOR
\RETURN $\ntc^*,\routes$
\end{algorithmic}
\end{algorithm}

Algorithm \ref{alg:network_improvement} is designed to be run in a continuous loop where the number of sampled configurations, $\texttt{max\_it}$, is selected to respect the real-time requirements of the system. It takes in the current network and task team configurations and the target network configuration from the previous iteration, $\ntc^*$, and returns an updated target network configuration, $\ntc^*$ (note that algorithm \ref{alg:network_improvement} is executed in a continuous loop and may run again before $\ntc$ reaches $\ntc^*$). In Steps 1 and 2, the set of optimal routing variables with respect to the current team configuration, $\routes$, and with respect to the current task team and optimal network team configuration, $\routes^*$, are computed by solving the robust routing optimization problem \eqref{eq:robust_routing} ($\texttt{SOCP}(\cdot)$ returns the routing variables of the associated solution). It is necessary to recompute $\routes^*$ as the task team may have moved since the last call even though the value of $\ntc^*$ remains the same between iterations of the algorithm. Step 3 finds the constraint closest to being violated for the current optimal configuration, which serves as the benchmark candidate configurations must beat in the local search performed in Steps 4-10. The function $\texttt{draw\_sample}(\ntc)$ returns a collision free ($||x_i-x_j|| > d_c$ for $i,j\in\{1,\ldots,n\}$ and $i\neq j$, given the safety margin $d_c>0$) network team configuration in the neighborhood of $\ntc$ drawn from some probability distribution. In our case, samples were drawn from a normal distribution centered at the current configuration. Then if the candidate configuration $\bs{x}_p$ is found to be better, the optimal configuration is updated and the search continues. Since $\ntc^*$ persists across iterations, Algorithm \ref{alg:network_improvement} drives the network team towards locally optimal network configurations.

\section{Numerical Results}\label{sec:num_results}

\subsection{Simulations}

\begin{figure}[t]
    \centering
    \subfigure[]{
  		\includegraphics[width=0.45\columnwidth]{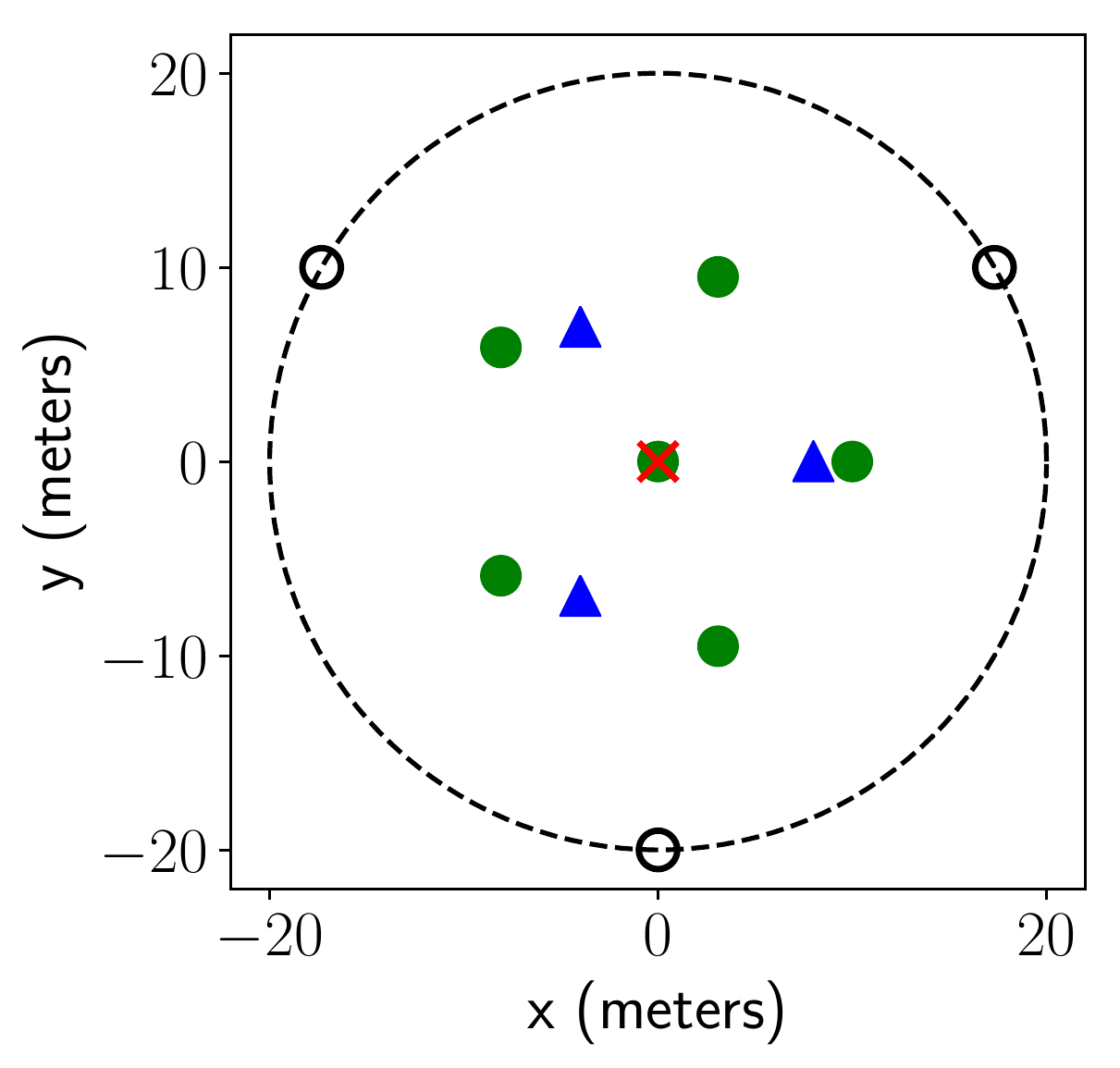}
     	\label{fig:sim_fixed_configs}
	}~
    \subfigure[]{
  		\includegraphics[width=0.45\columnwidth]{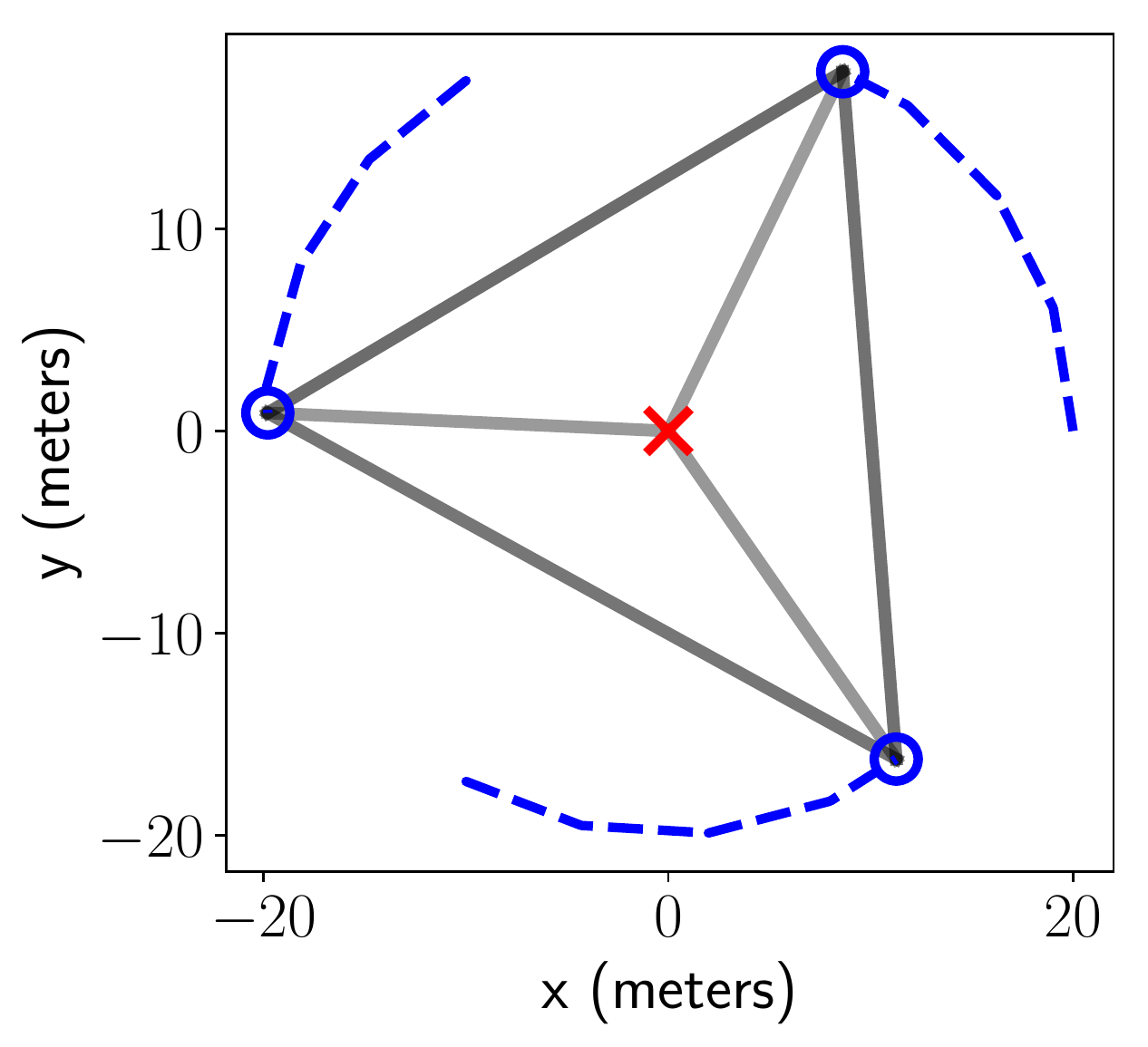}
  		\label{fig:sim_3_task_1_network}
	}\\
    \subfigure[]{
  		\includegraphics[width=0.45\columnwidth]{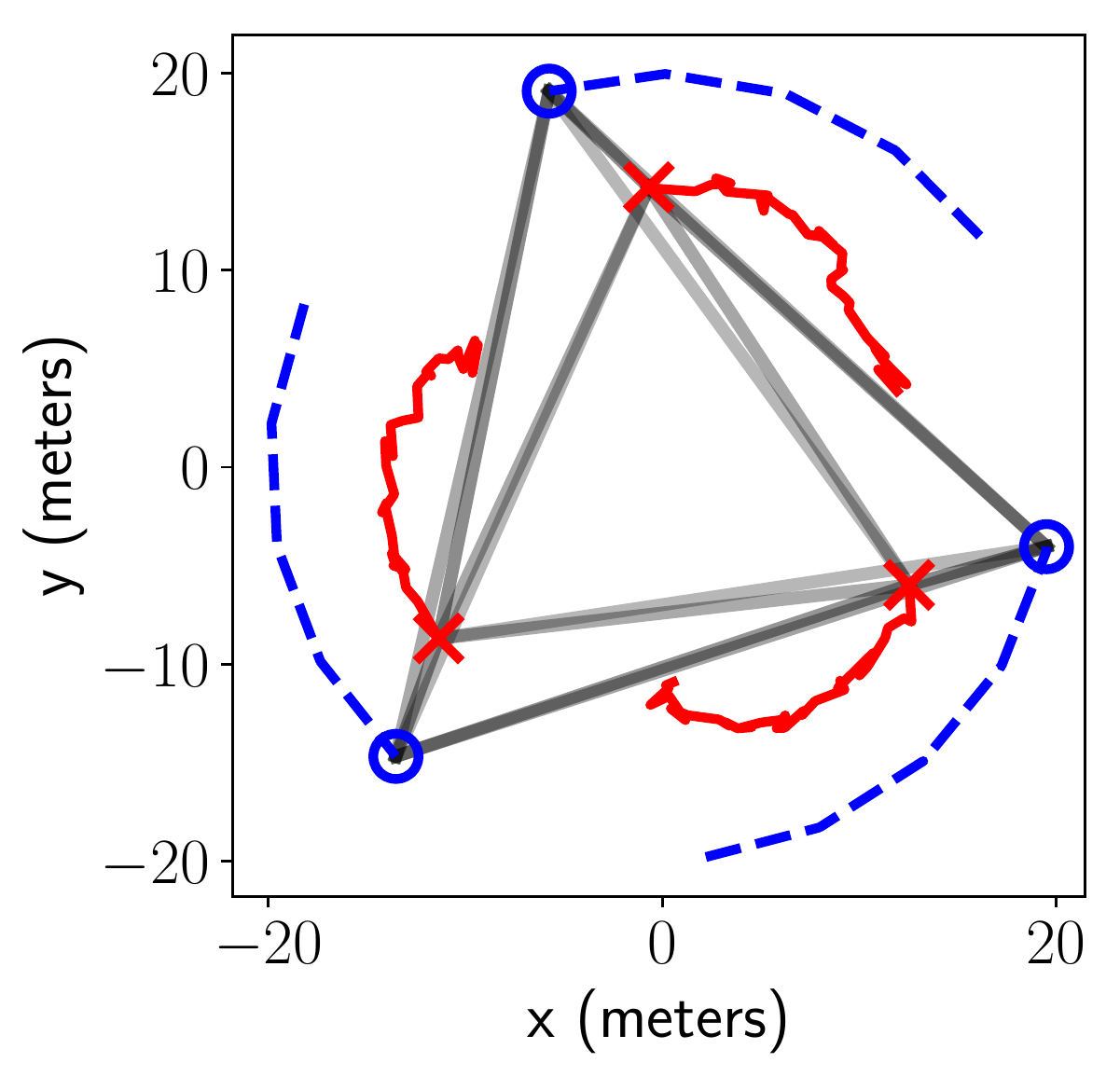} ~
  		\label{fig:sim_3_task_3_network}
	}~	 
	\subfigure[]{
  	    \includegraphics[width=0.45\columnwidth]{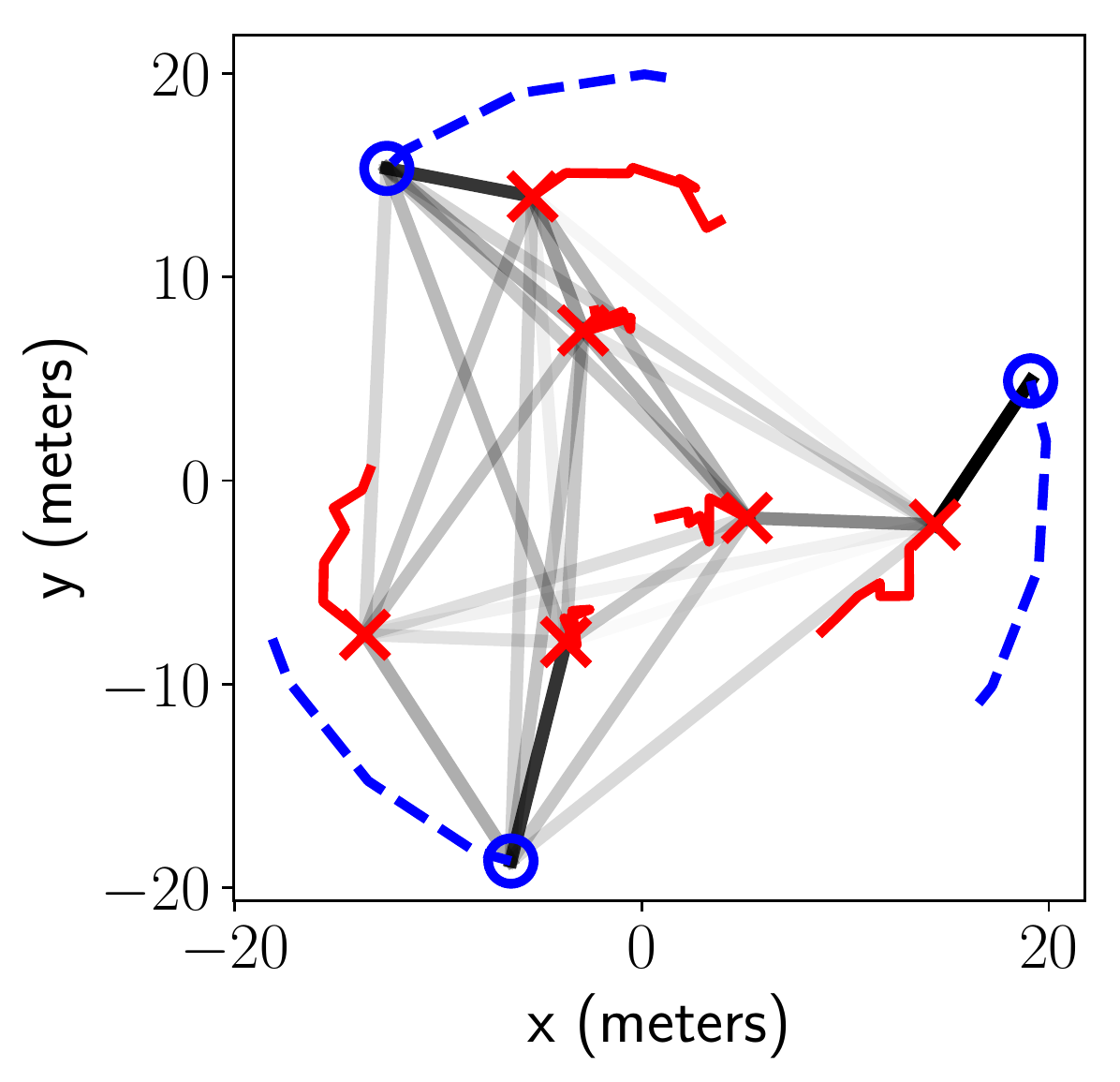} ~
  	    \label{fig:sim_3_task_6_network}
	}	 
	\caption{Representative team configurations for the $p=3$ task agent patrol with radius $20$m showing a) the fixed configurations used for comparison and mobile network teams comprised of b) $q=1$ c) $q=3$ and d) $q=6$ agents. Task agents are represented as blue circles, network agents as red crosses and their recent paths as dashed and solid lines, respectively. Grayscale lines connected the agents are network flows, with darker lines signifying links with higher usage.}
    \label{fig:sim_configurations}
\end{figure}


\begin{figure}[t!]
    \centering
    \subfigure[Resulting rates and $1-\epsilon$ confidence bound with $q=1$ network agent. Both the fixed and the dynamic team configurations are the same and cannot support the demanded rate specification.]{
  	\includegraphics[scale=1]{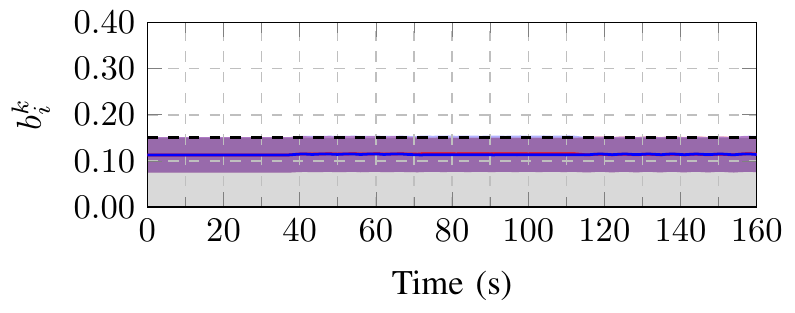}
     	\label{fig:3t1n}
	 }
    \subfigure[Resulting rates and $1-\epsilon$ confidence bound with $q=3$ network agents. The dynamic team successfully supports the demands while the fixed team sporadically fails to do so]{
  	\includegraphics[scale=1]{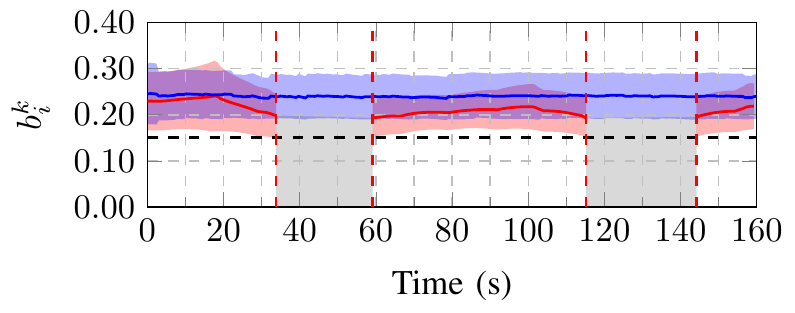}
  		\label{fig:3t3n}
	 }
    \subfigure[Resulting rates and $1-\epsilon$ confidence bound with $q=6$ network agents. Both the fixed configuration and the dynamic configuration can support the demand rate specification with the dynamic team performing better on average]{
  	\includegraphics[scale=1]{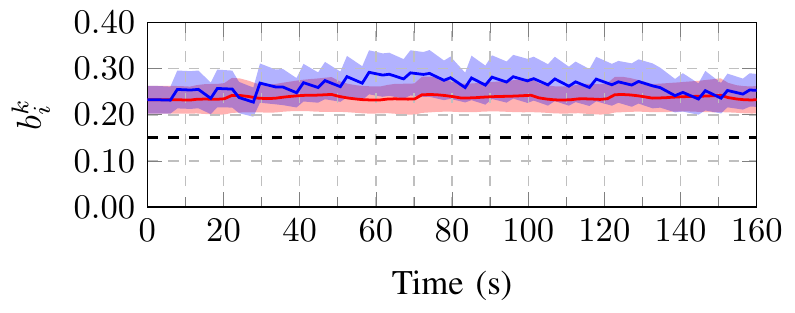}
  		\label{fig:3t6n}
	 }	 
	\caption{Average rate margin and confidence bound at the source node for $p=3$ task agents requiring a rate margin of $m_i^k=0.15$ with confidence $1-\epsilon=0.7$. The average rate margin of the fixed network team is shown in red and the dynamic team executing our local controller in blue. The $1-\epsilon$ confidence bound for each margin is shown as the shaded region.}
    \label{fig:rates}
\end{figure}

In order to verify our proposed algorithm, we implemented our system in ROS \cite{ros} and performed a set of simulations using Gazebo \cite{gazebo}. The state of each link, and in turn the network, was judged using the model described in equations \eqref{eq:channel_mean}, \eqref{eq:channel_var}. We considered a scenario where a team of three task agents patrolling a perimeter desired to exchange information but their relative distance precluded supporting communication via direct links. Thus, we deployed a team of network agents to act as relays, ensuring critical information from the patrolling agents was delivered to the others.

The communication requirements of the task team were modeled as three flows: $S_1=\{1\}$ with $D_1=\{2,3\}$; $S_2=\{2\}$ with $D_2=\{1,3\}$; and $S_3=\{3\}$ with $D_3=\{1,2\}$. For each flow the required margin was $m_i^k=0.15$ with confidence $1-\epsilon_k=0.7$. The channel parameters in question were chosen for IEEE 802.11 \cite{fink2011communication} with exact values listed in Fig. \ref{fig:rate}. Our system does not directly prescribe the number of agents needed to satisfy the task requirements. Thus, we conducted simulations with network teams of 1, 3, and 6 agents running our local controller. As a point of comparison, we also conducted the same simulations with a fixed network team deployed in a configuration to best cover the task space but not allowed to move with the task agents. The fixed network configurations as well as snapshots of the mobile network team in action are shown in the Fig. \ref{fig:sim_configurations}.

The performance of the network team in each scenario was measured as the average of the rate margins at the source nodes: $b_1^1$, $b_2^2$, and $b_3^3$. Since the optimization problem \eqref{eq:robust_routing} is only feasible if all flows are satisfied, this ensemble approach is an effective measure of system performance. The average rate margin of the fixed and dynamic teams are shown in Fig. \ref{fig:rates}.

Fig. \ref{fig:sim_3_task_1_network} shows the configuration for a single network agent. In this case, the fixed and dynamic configurations coincide at the center of the circle: regardless of initial position, our algorithm pulls the network agent to the optimal configuration at the center and evenly distributes traffic across the links. When $q=1$ the system cannot satisfy the task requirements of $m_i^k=0.15$ and $1-\epsilon=0.7$ and is only able to support a rate around $\sim0.12$ as shown in Fig. \ref{fig:3t1n}.

For three agents, the fixed configuration was chosen to be a triangle centered in the circle as seen in Fig. \ref{fig:sim_fixed_configs}. While this fixed triangle configuration performed well when the task team was aligned with it, the average rate margin suffered as the task team moved away and ultimately was unable to satisfy the communication requirements over the entire patrol trajectory as illustrated in the grey regions of Fig. \ref{fig:3t3n}. On the contrary, the three agent network team running our local controller converged to a triangle that rotated along with the patrol agents (Fig. \ref{fig:sim_3_task_3_network}) providing consistent communication throughout the duration of the task.

Finally, a deployment with six network agents is shown in Fig. \ref{fig:sim_3_task_6_network}. In this case the best fixed team configuration is not immediately obvious. We opted for a space filling pentagon with an agent at the center to cover the interior of the circle. With an increase in the number of agents their relative positioning becomes less crucial; while the dynamic network team performed better on average, the fixed team was also able to support the required rate over the duration of the patrol as evidenced in Fig. \ref{fig:3t6n}.

\subsection{Experiments}

We also conducted experiments to demonstrate the feasibility of our system on physical platforms using conventional IEEE 802.11n WiFi. For these tests we used the Intel Aero quadrotor research platform equipped with an Intel Atom x7-Z8750 processor, Intel Dual Band Wireless-AC 8260 Wi-Fi chip configured in IBSS (ad-hoc) mode, and Linux with ROS. Those familiar with conducting wireless system tests on physical platforms know the inherent difficulty of such an endeavor. In particular, our system 1) is based a probabilistic routing protocol and must be able to dynamically change the gateways (i.e. destination node) used for system packet routing 2) is centralized and must aggregating the state information of and disseminate routing updates to every agent in the network and 3) must gather gather meaningful network statistics without infringing upon 1,2. While a thorough treatment of each of these tasks falls beyond the scope of this paper, a summary of each module follows.

\begin{figure}[t]
    \centering
    \subfigure[Throughput]{
  	    \includegraphics[scale=1]{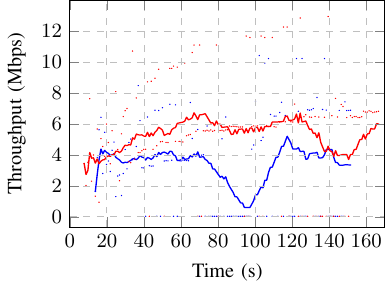}
     	\label{fig:experiment_throughput}
	}~
    \subfigure[Delay]{
  	    \includegraphics[scale=1]{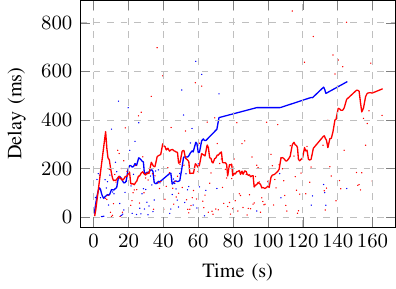}
  		\label{fig:experiment_delay}
	}
	\caption{A moving average of the a) delay and b) throughput for the live experiment. Statistics from the test using probabilistic routing are in red and from the test using a fixed, direct link are in blue. }
    \label{fig:experiment_statistics}
\end{figure}

The solution of problem \eqref{eq:robust_routing} is a matrix of routing variables, $\routes$, that define where data streams should be directed en route to their destination. Taking Fig. \ref{fig:exampleLowMarginHighConf} as an example, node 2 can send data through node 3 or directly to node 1. Instead of choosing a gateway for each incoming packet, we sample a fixed routing configuration from $\routes$ and enforce it for all packet transmissions over a chosen time window. Provided this time window is sufficiently small compared to the rate at which new versions of $\routes$ are received, this approximates the routing probabilities given by problem \eqref{eq:robust_routing}. On a Linux system, we accomplished this by adding and deleting routing tables at the network layer.

Our system is inherently centralized as both problem \eqref{eq:robust_routing} and algorithm \eqref{alg:network_improvement} require full state information. We arbitrarily nominate one of the quadrotors as the planning node and subsequently require that $\routes$ connect every agent do it. In practice this can be accomplished by enforcing additional flows for each agent to the planner node. Since the frequency of routing updates is low ($\sim1$s) we utilize the broadcast channel, which sends packets to all agents within reach, coupled with a network flooding rebroadcast policy to ensure every agent gets the latest copy of $\routes$.

Finally, for the task agent flows we utilized the iperf3 and traceroute Linux utilities to measure throughput and delay. iperf3 spams the network interface with packets to a given destination (just as a source node would) while traceroute measures the round trip time of packets. Since we are modifying the underlying routing table, any messages that get send through the network interface are subject to our routing protocol. The same applies to ROS messages.

\begin{figure}[t]
    \centering
  	\raisebox{-0.5\height}{\includegraphics[width=0.6\columnwidth]{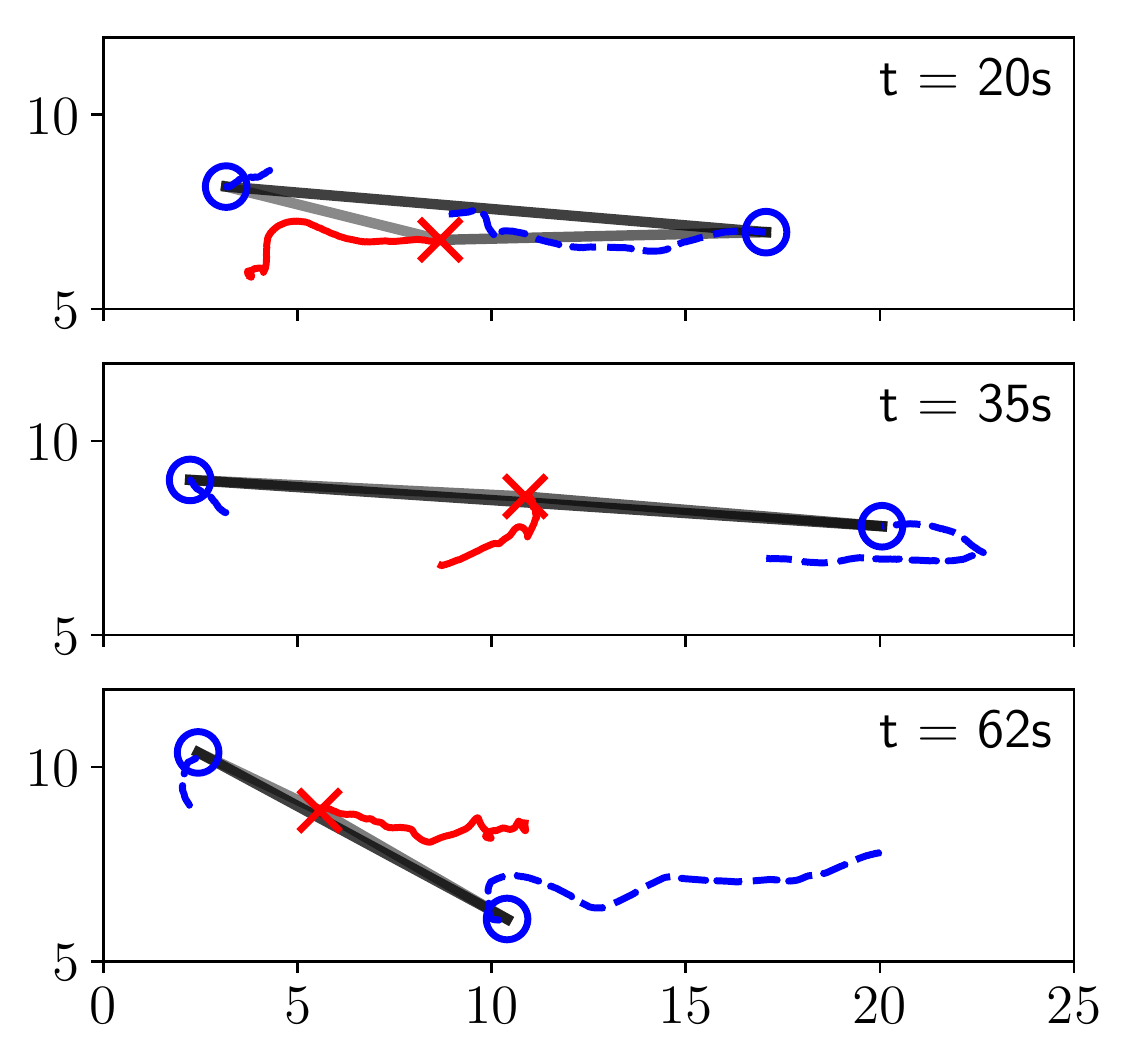}}
  	\hspace*{0.1cm}
  	\raisebox{-0.5\height}{\includegraphics[width=0.3\columnwidth]{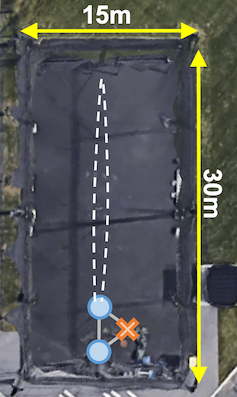}}
  	\caption{Snapshots of the agent paths and network routes at three different points during the line experiment. Task agents are represented as blue circles, the network agent as a red cross and their recent paths as dashed and solid lines, respectively. Grayscale lines connected the agents are network flows, with darker lines signifying links with higher usage. Units are in meters.}
    \label{fig:experiment_positions}
\end{figure}

Our experiment with three robots was conducted outdoors at the Penn Engineering Research and Collaboration Hub (PERCH) using one network agent maintaining a connection between two task agents, one acting as a base station and one free to roam. Fig. \ref{fig:experiment_statistics} shows the throughput and delay and Fig. \ref{fig:experiment_positions} shows the flight path of the robots during the test. We collected statistics for a flow from the roving task agent to the base station using 1) our robust routing protocol compared against 2) a fixed, direct link between source and destination not utilizing the intermediate network node. It is clear from Fig. \ref{fig:experiment_statistics} that as the roving task agent moves away from the base station the throughput drops and the delay grows for the case with direct routing; however, for the test using our probabilistic routing protocol the throughput remains constant even as the direct link is severed, demonstrating the utility of our system. Additionally, in Fig. \ref{fig:experiment_positions} our local controller keeps the network agent favorably positioned along the line connecting the two task agents.


\section{Conclusion}\label{sec:conclusions}

In this paper, we have introduced a task agnostic architecture for providing mobile wireless network infrastructure on demand. Network requirements defined through a task independent interface are satisfied by our system that jointly optimizes routing choices for packet transmission and network node placement. As a result, a task team can achieve their objective presuming on the availability of wireless communication.

\bibliographystyle{ieeetr} 
\bibliography{references}

\begin{thebibliography}{10}

\bibitem{kim2005maximizing}
Y.~Kim and M.~Mesbahi, ``On maximizing the second smallest eigenvalue of a
  state-dependent graph laplacian,'' in {\em Proceedings of the 2005, American
  Control Conference, 2005.}, pp.~99--103, IEEE, 2005.

\bibitem{stump2008connectivity}
E.~Stump, A.~Jadbabaie, and V.~Kumar, ``Connectivity management in mobile robot
  teams,'' in {\em 2008 IEEE international conference on robotics and
  automation}, pp.~1525--1530, IEEE, 2008.

\bibitem{zavlanos2007potential}
M.~M. Zavlanos and G.~J. Pappas, ``Potential fields for maintaining
  connectivity of mobile networks,'' {\em IEEE Transactions on robotics},
  vol.~23, no.~4, pp.~812--816, 2007.

\bibitem{zavlanos2008distributed}
M.~M. Zavlanos and G.~J. Pappas, ``Distributed connectivity control of mobile
  networks,'' {\em IEEE Transactions on Robotics}, vol.~24, no.~6,
  pp.~1416--1428, 2008.

\bibitem{ji2007distributed}
M.~Ji and M.~Egerstedt, ``Distributed coordination control of multiagent
  systems while preserving connectedness,'' {\em IEEE Transactions on
  Robotics}, vol.~23, no.~4, pp.~693--703, 2007.

\bibitem{de2006decentralized}
M.~C. De~Gennaro and A.~Jadbabaie, ``Decentralized control of connectivity for
  multi-agent systems,'' in {\em Proceedings of the 45th IEEE Conference on
  Decision and Control}, pp.~3628--3633, IEEE, 2006.

\bibitem{spanos2005motion}
D.~P. Spanos and R.~M. Murray, ``Motion planning with wireless network
  constraints,'' in {\em Proceedings of the 2005, American Control Conference,
  2005.}, pp.~87--92, IEEE, 2005.

\bibitem{notarstefano2006maintaining}
G.~Notarstefano, K.~Savla, F.~Bullo, and A.~Jadbabaie, ``Maintaining
  limited-range connectivity among second-order agents,'' in {\em 2006 American
  control conference}, pp.~6--pp, IEEE, 2006.

\bibitem{schuresko2009distributed}
M.~Schuresko and J.~Cort{\'e}s, ``Distributed motion constraints for algebraic
  connectivity of robotic networks,'' {\em Journal of Intelligent and Robotic
  Systems}, vol.~56, no.~1-2, pp.~99--126, 2009.

\bibitem{zavlanos2012network}
M.~M. Zavlanos, A.~Ribeiro, and G.~J. Pappas, ``Network integrity in mobile
  robotic networks,'' {\em IEEE Transactions on Automatic Control}, vol.~58,
  no.~1, pp.~3--18, 2012.

\bibitem{tekdas2010robotic}
O.~Tekdas, W.~Yang, and V.~Isler, ``Robotic routers: Algorithms and
  implementation,'' {\em The International Journal of Robotics Research},
  vol.~29, no.~1, pp.~110--126, 2010.

\bibitem{mostofi2008communication}
Y.~Mostofi, ``Communication-aware motion planning in fading environments,'' in
  {\em 2008 IEEE International Conference on Robotics and Automation},
  pp.~3169--3174, IEEE, 2008.

\bibitem{mostofi2009characterization}
Y.~Mostofi, A.~Gonzalez-Ruiz, A.~Gaffarkhah, and D.~Li, ``Characterization and
  modeling of wireless channels for networked robotic and control systems-a
  comprehensive overview,'' in {\em 2009 IEEE/RSJ International Conference on
  Intelligent Robots and Systems}, pp.~4849--4854, IEEE, 2009.

\bibitem{yan2012robotic}
Y.~Yan and Y.~Mostofi, ``Robotic router formation in realistic communication
  environments,'' {\em IEEE Transactions on Robotics}, vol.~28, no.~4,
  pp.~810--827, 2012.

\bibitem{fink2013robust}
J.~Fink, A.~Ribeiro, and V.~Kumar, ``Robust control of mobility and
  communications in autonomous robot teams,'' {\em IEEE Access}, vol.~1,
  pp.~290--309, 2013.

\bibitem{fink2011communication}
J.~Fink, ``Communication for teams of networked robots,'' 2011.

\bibitem{fink2011robust}
J.~Fink, A.~Ribeiro, and V.~Kumar, ``Robust control for mobility and wireless
  communication in cyber--physical systems with application to robot teams,''
  {\em Proceedings of the IEEE}, vol.~100, no.~1, pp.~164--178, 2011.

\bibitem{stephan2017concurrent}
J.~Stephan, J.~Fink, V.~Kumar, and A.~Ribeiro, ``Concurrent control of mobility
  and communication in multirobot systems,'' {\em IEEE Transactions on
  Robotics}, vol.~33, no.~5, pp.~1248--1254, 2017.

\bibitem{shorey2006mobile}
R.~Shorey, A.~Ananda, M.~C. Chan, and W.~T. Ooi, {\em Mobile, wireless, and
  sensor networks: technology, applications, and future directions}.
\newblock John Wiley \& Sons, 2006.

\bibitem{ros}
``{Robot Operating System (ROS)}.'' \url{https://www.ros.org/}.

\bibitem{gazebo}
``{Gazebo Simulator}.'' \url{http://gazebosim.org/}.

\end{thebibliography}

\end{document}